\documentclass[conference]{IEEEtran}

\usepackage{cite}
\usepackage{amsmath,amssymb}
\usepackage{graphicx}
\usepackage{booktabs}
\usepackage[ruled,vlined]{algorithm2e}
\usepackage[hidelinks]{hyperref}

\graphicspath{{figures/}}

\begin{document}

\title{Control Barrier Functions for Safety-Aware\\Latent-Space Motion Planning}

\author{\IEEEauthorblockN{Chandrakiran Ravula and Siddhanth Ramesh}
\IEEEauthorblockA{Indian Institute of Science (IISc), Bengaluru, India}}

\maketitle

\begin{abstract}
Latent-space path planning (LSPP) compresses the configuration space of a robotic manipulator into a compact, differentiable manifold learned by a variational autoencoder, enabling fast gradient-based planning without explicit inverse kinematics. Existing LSPP frameworks enforce obstacle avoidance with a binary collision classifier, whose saturated gradients near obstacle boundaries and lack of temporal structure cause the optimizer to stall and provide no forward-invariance guarantee. We replace the classifier with a neural control barrier function (CBF) learned directly in the latent space. The barrier is trained with a three-term objective that combines safe and unsafe sign constraints with a discrete-time decrease condition evaluated on transitions of an obstacle-blind nominal planner, and it is enforced at planning time by a closed-form, first-order correction that projects each goal-directed latent step back onto the CBF constraint. On a 7-DoF Franka Panda manipulator, the CBF planner achieves a 75.6\% success rate on a fixed-obstacle task, 12.0 percentage points above the best of 35 tuned classifier configurations, and 83.8\% on a multi-position task. A sweep of the decay rate $\alpha$ characterizes the safety--performance trade-off, with the best operating points in $\alpha \in [0.10, 0.20]$. In a fully general random-obstacle setting, success drops to 33.0\%, which we trace to covariate shift between the obstacle-blind training transitions and the states visited by the corrected planner.
\end{abstract}


\section{Introduction}

\begin{figure}[t]
    \centering
    \includegraphics[width=\columnwidth]{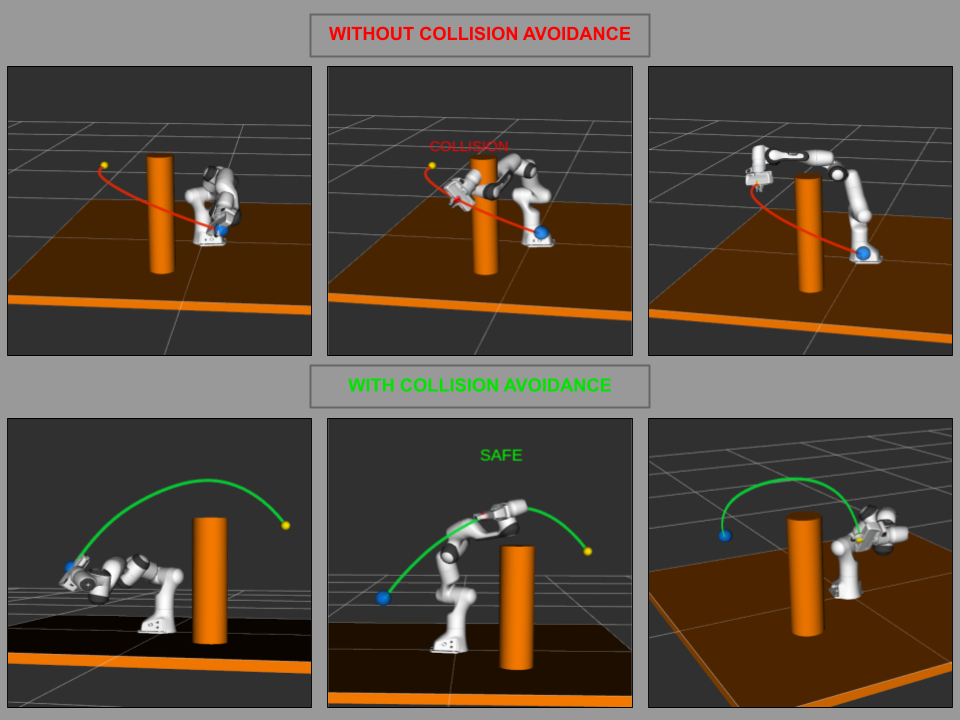}
    \caption{\textbf{Top:} A naive trajectory executed without collision avoidance intersects the obstacle. \textbf{Bottom:} A safe trajectory guides the manipulator over the obstacle to the target.}
    \label{fig:keyframe_concept}
\end{figure}

Motion planning for robotic manipulators in high-dimensional configuration spaces remains computationally demanding. A 7-degree-of-freedom (DoF) manipulator such as the Franka Emika Panda~\cite{gallouedec2021pandagym} must reach a goal while satisfying joint limits, avoiding workspace obstacles, and remaining kinematically feasible (Fig.~\ref{fig:keyframe_concept}). Safe and reliable operation is essential in safety-critical domains such as healthcare~\cite{gottesman2019guidelines} and bipedal locomotion~\cite{nguyen2015safety}. Sampling-based planners such as Rapidly-exploring Random Trees (RRT) and Probabilistic Roadmaps (PRM) are probabilistically complete~\cite{karaman2011sampling}, but their runtime scales poorly with dimension. Optimization-based planners such as CHOMP~\cite{ratliff2009chomp} and TrajOpt~\cite{schulman2013finding} scale better, but they are sensitive to initialization, prone to local minima, and rely on expensive distance-field queries or exact collision checks at every iteration. Model predictive control (MPC)~\cite{MAYNE2000789} offers a principled receding-horizon alternative but faces the same dimensionality.

Learned latent representations offer an alternative by compressing the planning problem into a lower-dimensional, smooth manifold. A variational autoencoder (VAE)~\cite{kingma2014auto, higgins2017beta} trained on valid robot configurations learns a compact latent space that implicitly captures kinematic feasibility and the multimodality of inverse kinematics. Gradient-based optimization in this space, termed activation maximization (AM), can then pursue goal reaching and obstacle avoidance jointly. Hung et al.~\cite{hung2022reaching} instantiate this idea for manipulation, building on the broader paradigm of planning in learned generative manifolds~\cite{ichter2019robot}: a VAE encodes joint configurations and end-effector positions, and a learned binary collision classifier steers the latent state away from predicted collisions.

Relying on a binary classifier for obstacle avoidance has two structural weaknesses. First, its sigmoid output yields a poor gradient landscape: far from an obstacle the gradient is nearly zero and provides no anticipatory guidance, while near or inside the obstacle it spikes. This ``cliff-edge'' profile traps the trajectory optimizer in local minima. Second, a classifier evaluates static configurations and has no temporal structure, so nothing prevents a state judged safe at step $k$ from moving into collision at step $k+1$.

We address both weaknesses by learning a control barrier function (CBF)~\cite{ames2014control, Ames2019CBF} directly in the VAE latent space. A CBF $B(z)$ is a continuous function whose nonnegative superlevel set is the safe set; its decrease condition bounds how fast $B$ may decay from one step to the next, which renders the safe set forward invariant. Rather than adding a soft collision penalty to the planning objective, we use the learned barrier as a closed-form safety filter: each goal-directed latent step is computed without regard to obstacles and, if it violates the CBF condition, is projected along the barrier gradient back onto the constraint. Neural CBFs have emerged as a data-driven way to synthesize such certificates~\cite{dawson2023safe}, but their integration into learned generative planning spaces remains underexplored. Our contributions are:
\begin{itemize}
    \item A reproduction of the LSPP baseline~\cite{hung2022reaching} for a 7-DoF Franka Panda, with a two-stage evaluation that validates every plan by exact geometric collision checking.
    \item A neural CBF defined in the learned latent space and trained with a three-term objective (safe sign, unsafe sign, and discrete decrease condition) on a state-label dataset and on transitions of an obstacle-blind nominal planner.
    \item A closed-form, first-order safety correction that filters each latent planning step through the CBF constraint, decoupling safety from goal pursuit.
    \item An evaluation of three barrier variants of increasing generality, $B(z)$, $B(z, x, y)$, and $B(z, o)$, with a systematic characterization of the $\alpha$ safety--performance trade-off. The CBF outperforms the best-tuned classifier by 12.0 percentage points on a fixed-obstacle task and reaches 83.8\% success on a multi-position task; in the general setting we identify covariate shift as the main failure mode.
\end{itemize}

Section~\ref{sec:related} reviews related work and Section~\ref{sec:background} presents the background and baseline framework. Section~\ref{sec:method} introduces the latent-space CBF, Sections~\ref{sec:setup} and~\ref{sec:results} describe the experimental setup and results, and Sections~\ref{sec:limitations} and~\ref{sec:conclusion} discuss limitations, future work, and conclusions.

\section{Related Work}\label{sec:related}

\subsection{Motion Planning and Learned Representations}

Sampling-based algorithms avoid constructing the configuration space explicitly, but in high dimensions they yield highly variable planning times and jerky paths that require post-processing. Trajectory optimization minimizes a cost functional combining smoothness and obstacle avoidance~\cite{ratliff2009chomp, schulman2013finding}, but complex geometry is hard to represent analytically and non-convex environments trap the optimizer. Both paradigms rely on forward kinematics and geometric intersection tests inside the planning loop. Learning from demonstrations offers a data-driven alternative, including behavioral cloning~\cite{pomerleau1991efficient}, inverse reinforcement learning~\cite{ng2000algorithms}, generative adversarial imitation~\cite{ho2016generative}, and goal-conditioned imitation that generalizes across targets~\cite{ding2020gcil}. Generative models enable planning in learned manifolds~\cite{ichter2019robot, hung2022reaching}, and diffusion models have been used directly as trajectory planners~\cite{janner2022planning}. We keep the latent planner of~\cite{hung2022reaching} and replace its safety mechanism.

\subsection{Control Barrier Functions and Neural Certificates}

CBFs were introduced in the context of adaptive cruise control~\cite{ames2014control} and formalized in~\cite{ames2017control, Ames2019CBF}. They have since been applied to UAV flight~\cite{barry2012safety, tayal2024control}, bipedal walking~\cite{nguyen2015safety}, and quadrotor control~\cite{dawson2022safe}. When analytical barriers are unavailable, they can be learned from data: Robey et al.~\cite{robey2020learning} learn barrier certificates from expert demonstrations, and Srinivasan et al.~\cite{srinivasan2020synthesis} enforce control-theoretic inequalities as empirical losses over sampled states. Dawson et al.~\cite{dawson2022safe, dawson2023safe} learn neural Lyapunov-barrier functions with formal guarantees. Exact verification of ReLU-based neural CBFs~\cite{NEURIPS2023_120ed726}, automated certificate synthesis with FOSSIL~\cite{abate2021fossil}, training under input constraints~\cite{so2024train}, and conformal prediction for verified neural CBFs~\cite{tayal2025cp} address the reliability of learned certificates, and physics-informed objectives can improve them further~\cite{tayal2025physics}. Hamilton--Jacobi (HJ) reachability~\cite{bansal2017hamilton} characterizes backward reachable sets exactly; Fisac et al.~\cite{Fisac2019HJSafety} connect HJ safety analysis with reinforcement learning, and robust control barrier-value functions~\cite{choi2021robust} combine the CBF and HJ perspectives.

\subsection{Safe Learning from Data}

Surveys of safe reinforcement learning (RL)~\cite{garcia2015comprehensive, Zhao2023SafeRL} group approaches into constrained optimization~\cite{altman2021constrained, achiam2017constrained}, shielding~\cite{alshiekh2018safe}, and barrier-based methods, building on standard RL foundations~\cite{sutton1998introduction}. Offline RL learns from fixed datasets~\cite{levine2020OffRL}, where extrapolation error~\cite{fujimoto2019off} and bootstrapping error~\cite{kumar2019stabilizing} motivate conservative~\cite{kumar2020CQL} and implicit~\cite{kostrikov2022offline} value learning. In offline safe RL, feasibility-guided diffusion~\cite{zheng2024fisor}, reachability-based flow policies~\cite{tayal2026safe}, and epigraph-guided flow matching~\cite{tayal2026epiflow} show that expressive generative policies can balance safety and task performance, while value-guided offline CBFs learn safety filters entirely from offline data~\cite{tayal2025vocbf}. Safety has also been built into imitation learning through end-to-end imitation with CBF guarantees~\cite{cosner2022end}, in-distribution barrier functions~\cite{Fernando2023iDBF}, robust imitation through stochastic encoding~\cite{tayal2025rise}, safety-guided imitation~\cite{ciftci2024safe}, soft barrier certificates for unknown stochastic environments~\cite{wang2023enforcing}, and semi-supervised safe visuomotor policies~\cite{tayal2024semi}.

\section{Background and Baseline Framework}\label{sec:background}

\subsection{Problem Setting}

Given a start configuration $q_{\mathrm{start}}$ and a target end-effector position $e_{\mathrm{target}}$, the planner must find a continuous trajectory $\tau(t) \in \mathcal{C}_{\mathrm{free}}$ that ends at a configuration reaching $e_{\mathrm{target}}$, where $\mathcal{C}_{\mathrm{free}}$ is the collision-free subset of the configuration space. Following~\cite{hung2022reaching}, the state of the Franka Panda is $x = (q, e) \in \mathbb{R}^{10}$, where $q \in \mathbb{R}^7$ are the joint angles and $e \in \mathbb{R}^3$ is the end-effector position given by forward kinematics. Including $e$ in the state lets the planner target positions directly without solving inverse kinematics; the relationship between $q$ and $e$ is learned implicitly. All ten dimensions are standardized with training-set statistics, and the collision classifier reuses the statistics of the free-space dataset so that it operates in the same normalized space as the VAE. Obstacles are cylinders described by $o = [x, y, h, r]$, their planar position, height, and radius.

\subsection{Latent Representation}

\begin{figure}[t]
    \centering
    \includegraphics[width=\columnwidth]{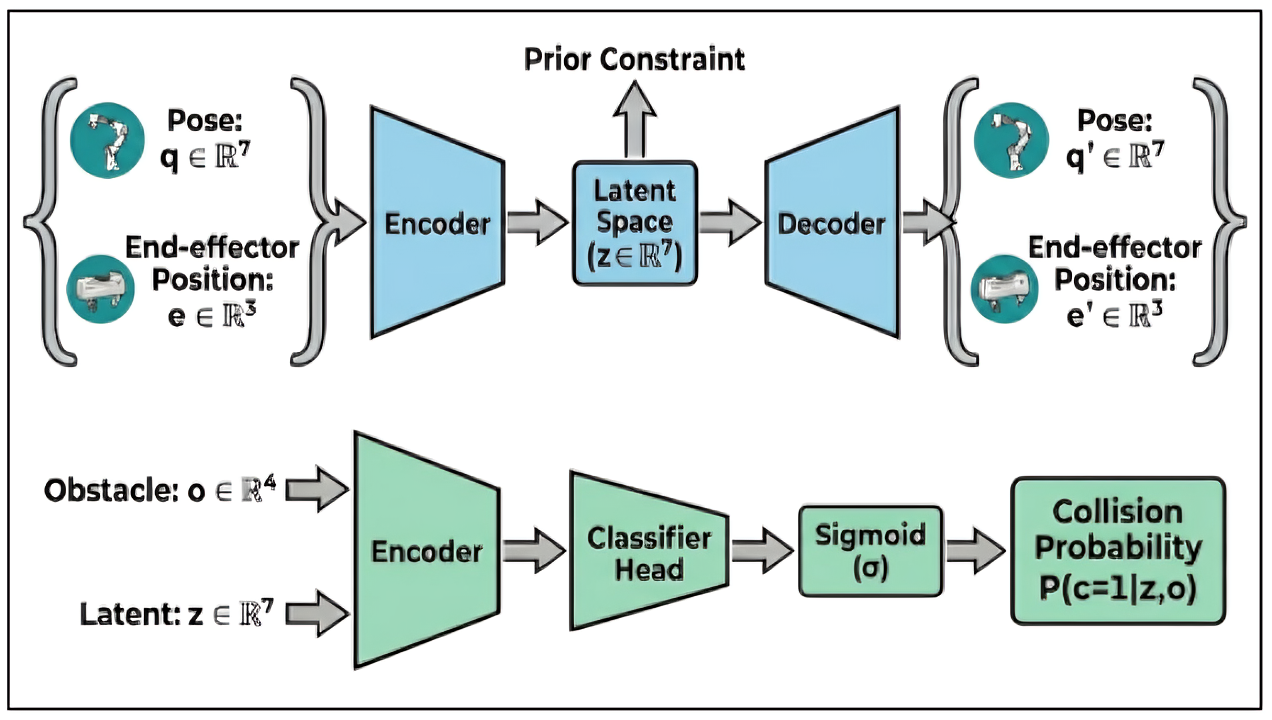}
    \caption{Baseline LSPP architecture. \textbf{Top:} The VAE compresses the robot state ($q \in \mathbb{R}^7$, $e \in \mathbb{R}^3$) into a 7-D latent space regularized by a prior. \textbf{Bottom:} The binary collision classifier predicts collision from the latent code concatenated with the obstacle descriptor $o \in \mathbb{R}^4$.}
    \label{fig:baseline_architecture}
\end{figure}

The VAE maps $x \in \mathbb{R}^{10}$ to a latent code $z \in \mathbb{R}^7$ (Fig.~\ref{fig:baseline_architecture}). The encoder has four hidden layers of 2048 units with exponential linear unit (ELU) activations and outputs $\mu \in \mathbb{R}^7$ and $\sigma^2 \in \mathbb{R}^7$, with $\sigma^2 = \mathrm{softplus}(\cdot) + 10^{-5}$ for numerical stability; samples are drawn as $z = \mu + \epsilon \odot \sigma$ with $\epsilon \sim \mathcal{N}(0, I)$. The decoder mirrors the encoder and outputs $\hat{x} = (\hat{q}, \hat{e})$.

The standard evidence lower bound (ELBO)~\cite{kingma2014auto} balances reconstruction against a Kullback--Leibler (KL) penalty toward the prior $p(z) = \mathcal{N}(0, I)$, and $\beta$-VAE~\cite{higgins2017beta} modulates this weight to obtain more disentangled representations. Neither guarantees the strict reconstruction fidelity that robot kinematics requires, so we use the Generalized ELBO with Constrained Optimization (GECO)~\cite{rezende2018taming}, which treats reconstruction quality as a constraint and minimizes
\begin{multline}\label{eq:geco}
\mathcal{L}_{\mathrm{GECO}} = D_{\mathrm{KL}}\big[q_\phi(z \mid x) \,\|\, p(z)\big] \\
+ \lambda\, \mathbb{E}_{z}\big[\max\big(0,\ \mathrm{MSE}(x, \hat{x}) - g_{\mathrm{goal}}\big)\big],
\end{multline}
where the Lagrange multiplier $\lambda$ is updated by an exponential moving average: it grows while the mean squared error exceeds the threshold $g_{\mathrm{goal}}$ and shrinks once the constraint is met, letting the network prioritize compression. We train for 16{,}000 epochs with Adam (learning rate $10^{-4}$, batch size 4096) and $g_{\mathrm{goal}} = 0.0008$. The checkpoint is selected by the area under the curve of an AM planning metric rather than by validation loss, so that the chosen model supports stable gradient-based planning.

\subsection{Collision Classifier}

In the baseline, a classifier head is appended to the frozen VAE encoder. It takes the 11-D input $(z, o)$, has four hidden layers of 2048 units, and outputs a logit $\ell(z, o)$ whose sigmoid gives $P(\mathrm{collision} \mid z, o)$. It is trained with binary cross-entropy (BCE) for 16{,}000 epochs.

\subsection{Activation Maximization Planning}

AM performs gradient descent on $z$ to minimize
\begin{align}
\mathcal{L}_{\mathrm{AM}}(z) &= \|\hat{e}(z) - e_{\mathrm{target}}\|^2 + \tfrac{\lambda_{\mathrm{prior}}}{2}\|z\|^2 \nonumber\\
&\quad - \lambda_{\mathrm{obs}} \sum_i \log\big(1 - \sigma(\ell(z, o_i)/T)\big). \label{eq:am}
\end{align}
The first term drives the end effector toward the target, the second keeps $z$ near the VAE prior to avoid out-of-distribution, kinematically invalid states, and the third minimizes collision probability, with temperature $T = 3.0$ softening the classifier gradients. The optimizer is initialized with a sample of the encoder distribution at $x_{\mathrm{start}}$ rather than its mean, which provides exploration noise in the non-convex loss landscape. Planning runs Adam with learning rate 0.15 for at most 300 steps; each iterate is decoded into a waypoint $(\hat{q}_k, \hat{e}_k)$, and planning stops once $\|\hat{e}_k - e_{\mathrm{target}}\| < 10$\,mm.

\subsection{Control Barrier Functions}

For a system $\dot{x} = f(x, u)$, the safe set $\mathcal{C} = \{x : B(x) \geq 0\}$ is the superlevel set of a continuously differentiable function $B$~\cite{ames2017control, Ames2019CBF}. $\mathcal{C}$ is forward invariant if there exists an extended class-$\mathcal{K}$ function $\alpha$ such that
\begin{equation}\label{eq:cbf_continuous}
\dot{B}(x) + \alpha(B(x)) \geq 0.
\end{equation}
For discrete iterates with step $\Delta t$ and a scalar decay rate $\alpha$ satisfying $0 \leq \alpha \Delta t \leq 1$, the corresponding condition is
\begin{equation}\label{eq:cbf_discrete}
B(z_{k+1}) \geq (1 - \alpha \Delta t)\, B(z_k),
\end{equation}
which bounds the fraction of the safety margin that may be lost in a single step.

\section{Latent-Space Control Barrier Functions}\label{sec:method}

We replace the classifier term of~\eqref{eq:am} with a neural barrier $B(z, o)$ trained in the latent space and enforced through a predictor--corrector step. The design draws on neural CBF synthesis~\cite{dawson2022safe, dawson2023safe, so2024train}, safety-guided imitation~\cite{ciftci2024safe, cosner2022end}, and in-distribution barrier functions~\cite{Fernando2023iDBF}.

\subsection{Why a Barrier Instead of a Classifier}

\textit{Gradient quality.} The BCE objective saturates for confident predictions. A barrier trained as an unbounded, signed function instead provides dense, non-vanishing gradients $\nabla_z B(z, o)$ that reflect the local boundary geometry, an advantage also observed for CBF-based obstacle avoidance on UAVs~\cite{tayal2024control}.

\textit{Temporal structure.} The discrete condition~\eqref{eq:cbf_discrete} couples $z_k$ and $z_{k+1}$ and rules out latent steps that remove most of the safety margin in a single iteration.

\textit{Modularity.} In the baseline, goal seeking and safety compete within one scalar loss. With a CBF, the nominal goal-directed step is computed freely and safety is applied afterwards as a projection, which avoids gradient interference. This mirrors the separation of policy and safety filter in safe visuomotor control~\cite{tayal2024semi}.

\subsection{Training Data}

The barrier must capture both spatial geometry (where obstacles are) and the planner's dynamics (how latent iterates move). Analytical obstacle boundaries are intractable in the non-linear 7-D latent space, so we learn the barrier from two datasets (Algorithm~\ref{alg:cbf_data_gen}):
\begin{itemize}
    \item \textbf{State-label dataset} $\mathcal{D}_{\mathrm{state}}$: collision-labeled configurations encoded deterministically through the VAE mean into tuples $(z, o, \mathrm{label})$, augmented with samples near obstacle surfaces to sharpen the decision boundary.
    \item \textbf{Transition dataset} $\mathcal{D}_{\mathrm{trans}}$: pairs $(z_k, z_{\mathrm{nom}})$ produced by a nominal planner that optimizes only the goal and prior terms and ignores obstacles, with each state labeled by a geometric simulator. Because the barrier will correct exactly this obstacle-blind optimizer at inference time, it must learn the decrease condition on that optimizer's transitions.
\end{itemize}
This follows Robey et al.~\cite{robey2020learning}, who learn barrier certificates from sampled trajectories when analytical models are unavailable, and Srinivasan et al.~\cite{srinivasan2020synthesis}, who enforce control-theoretic inequalities as empirical losses; here, the nominal planner's transitions play the role of expert trajectories.

\begin{algorithm}[t]
\caption{Dual-dataset generation for CBF training}\label{alg:cbf_data_gen}
\KwIn{Collision dataset $\mathcal{D}_{\mathrm{base}}$, nominal planner, geometric simulator}
\KwOut{$\mathcal{D}_{\mathrm{state}}$, $\mathcal{D}_{\mathrm{trans}}$}
$\mathcal{D}_{\mathrm{state}} \leftarrow \emptyset$, $\mathcal{D}_{\mathrm{trans}} \leftarrow \emptyset$\;
\ForEach{$(q, o, \mathrm{label}) \in \mathcal{D}_{\mathrm{base}}$}{
    $\mu, \sigma \leftarrow \mathrm{Encoder}(q)$\;
    $\mathcal{D}_{\mathrm{state}} \leftarrow \mathcal{D}_{\mathrm{state}} \cup \{(\mu, o, \mathrm{label})\}$\;
}
\For{$i = 1$ \KwTo $N_{\mathrm{scenarios}}$}{
    Sample $z_0 \sim \mathrm{Encoder}(q_{\mathrm{start}})$\;
    \For{$k = 0$ \KwTo $K_{\max}$}{
        \tcp{obstacle-blind nominal step}
        $z_{\mathrm{nom}} \leftarrow z_k - \mathrm{Adam}(\nabla(\mathcal{L}_{\mathrm{goal}} + \mathcal{L}_{\mathrm{prior}}))$\;
        $l_k \leftarrow \mathrm{Simulator.safe}(z_k)$\;
        $l_{\mathrm{nom}} \leftarrow \mathrm{Simulator.safe}(z_{\mathrm{nom}})$\;
        $\mathcal{D}_{\mathrm{trans}} \leftarrow \mathcal{D}_{\mathrm{trans}} \cup \{(z_k, z_{\mathrm{nom}}, o, l_k, l_{\mathrm{nom}})\}$\;
        $z_{k+1} \leftarrow z_{\mathrm{nom}}$\;
    }
}
\Return{$\mathcal{D}_{\mathrm{state}}, \mathcal{D}_{\mathrm{trans}}$}
\end{algorithm}

\subsection{Barrier Network and Training Objective}

The barrier network takes the 11-D input $(z, o)$, has four hidden layers of 2048 ELU units, and outputs a scalar $B \in \mathbb{R}$ without a final sigmoid; $B \geq 0$ denotes the safe set. It is trained to minimize
\begin{align}
\mathcal{L}_{\mathrm{CBF}} &= \lambda_s\, \mathbb{E}_{\mathrm{safe}}\big[\mathrm{ReLU}(\gamma - B(z, o))\big] \nonumber\\
&\quad + \lambda_u\, \mathbb{E}_{\mathrm{unsafe}}\big[\mathrm{ReLU}(B(z, o) + \gamma)\big] \nonumber\\
&\quad + \lambda_d\, \mathbb{E}_{\mathrm{trans}}\big[\mathrm{ReLU}(B_{\mathrm{target}} - B(z_{\mathrm{nom}}, o))\big], \label{eq:cbf_loss}
\end{align}
where $B_{\mathrm{target}} = (1 - \alpha \Delta t) B(z_k, o)$. The first two terms are evaluated on $\mathcal{D}_{\mathrm{state}}$ and the third on $\mathcal{D}_{\mathrm{trans}}$. The margin $\gamma = 1$ requires $B \geq 1$ on safe states and $B \leq -1$ on unsafe states; this two-unit gap buffers against VAE decoding variance. To isolate the difficulty of obstacle generalization, we train three variants:
\begin{itemize}
    \item \textbf{FixedBarrierNet} $B(z)$: 7-D input, one fixed obstacle.
    \item \textbf{MultiPosBarrierNet} $B(z, x, y)$: 9-D input, conditioned on the obstacle's planar position.
    \item \textbf{BarrierNet} $B(z, o)$: 11-D input, conditioned on fully random obstacle parameters.
\end{itemize}

\subsection{Closed-Form Safety Correction}

At planning time the obstacle term of~\eqref{eq:am} is removed and replaced by a predictor--corrector step (Algorithm~\ref{alg:cbf_correction}). The predictor takes a nominal Adam step $z_{\mathrm{nom}}$ on $\mathcal{L}_{\mathrm{goal}} + \mathcal{L}_{\mathrm{prior}}$. If $B(z_{\mathrm{nom}}, o) \geq B_{\mathrm{target}}$, the step is accepted; otherwise the corrector moves along the barrier gradient,
\begin{equation}\label{eq:correction}
z_{\mathrm{safe}} = z_{\mathrm{nom}} + \kappa\, \nabla_z B(z_{\mathrm{nom}}, o).
\end{equation}
A first-order Taylor expansion around $z_{\mathrm{nom}}$ gives
\begin{equation}
B(z_{\mathrm{nom}} + \kappa \nabla_z B) \approx B(z_{\mathrm{nom}}) + \kappa \|\nabla_z B(z_{\mathrm{nom}})\|^2,
\end{equation}
and setting the right-hand side equal to $B_{\mathrm{target}}$ yields the smallest step along the gradient that restores the constraint to first order,
\begin{equation}\label{eq:kappa}
\kappa = \max\left(0,\ \frac{B_{\mathrm{target}} - B(z_{\mathrm{nom}}, o)}{\|\nabla_z B(z_{\mathrm{nom}}, o)\|^2 + \epsilon}\right),
\end{equation}
which we clip to $\kappa \leq 1$ to bound the correction magnitude.

\textit{Optimizer state reset.} When the correction overwrites $z$, the Adam moment estimates no longer match the new iterate, and the accumulated momentum would push back against the projection in subsequent steps. We therefore reset Adam's internal state for $z$ whenever $\kappa > 0$.

\begin{algorithm}[t]
\caption{Closed-form CBF safety correction}\label{alg:cbf_correction}
\KwIn{Latent state $z_k$, obstacle $o$, decay rate $\alpha$, step $\Delta t$, barrier $B$}
\KwOut{Safe latent state $z_{\mathrm{safe}}$}
\tcp{predictor: nominal goal-directed step}
$z_{\mathrm{nom}} \leftarrow z_k - \mathrm{Adam}(\nabla(\mathcal{L}_{\mathrm{goal}} + \mathcal{L}_{\mathrm{prior}}))$\;
$B_{\mathrm{target}} \leftarrow (1 - \alpha \Delta t)\, B(z_k, o)$\;
\eIf{$B(z_{\mathrm{nom}}, o) \geq B_{\mathrm{target}}$}{
    $z_{\mathrm{safe}} \leftarrow z_{\mathrm{nom}}$\;
}{
    \tcp{corrector: project along the barrier gradient}
    $g \leftarrow \nabla_z B(z_{\mathrm{nom}}, o)$\;
    $\kappa \leftarrow \min\big(1,\ \max\big(0, \frac{B_{\mathrm{target}} - B(z_{\mathrm{nom}}, o)}{\|g\|^2 + \epsilon}\big)\big)$\;
    $z_{\mathrm{safe}} \leftarrow z_{\mathrm{nom}} + \kappa\, g$\;
    Reset Adam state for $z$\;
}
\Return{$z_{\mathrm{safe}}$}
\end{algorithm}

\subsection{Role of the Decay Rate}

The decay rate $\alpha$ sets the safety--performance trade-off. As $\alpha \to 0$, condition~\eqref{eq:cbf_discrete} becomes $B(z_{k+1}) \geq B(z_k)$ and forbids any decrease of the barrier, producing an overly conservative planner that stalls near obstacles. As $\alpha \Delta t \to 1$, the condition reduces to $B(z_{\mathrm{nom}}) \geq 0$, which permits large latent steps that violate the local linearity behind~\eqref{eq:kappa} and lead to collisions after decoding. Best performance is therefore expected at intermediate values, which we characterize empirically in Section~\ref{sec:results}.

\section{Experimental Setup}\label{sec:setup}

\subsection{Research Questions}

The evaluation addresses three questions: \textbf{(Q1)} Does the closed-form CBF correction outperform a tuned binary classifier under identical planning conditions? \textbf{(Q2)} How does $\alpha$ mediate the safety--performance trade-off? \textbf{(Q3)} How does conditioning the barrier on increasingly general obstacle parameters affect generalization?

\subsection{Environment, Data, and Evaluation Protocol}

The workspace respects the joint limits of the Franka Panda (e.g., $q_1 \in [-166^\circ, 166^\circ]$). The training corpus contains 100{,}000 free-space transitions and 100{,}000 obstacle collision states sampled uniformly within $x \in [0.4, 0.8]$\,m and $y \in [-0.5, 0.5]$\,m. Every method is evaluated with the same two-stage protocol. In Stage~1, the planner produces a latent trajectory using its learned safety signal. In Stage~2, the decoded joint trajectory is validated in an independent geometric simulator (Robo3D) that models robot links as capsules and computes exact minimum distances to cylindrical obstacles. A plan succeeds only if the end effector ends within 10\,mm of the target and no waypoint intersects an obstacle, so learned models never certify their own success.

\subsection{Experimental Settings}

Each of the three settings adds a degree of freedom to the obstacle parameterization:
\begin{itemize}
    \item \textbf{S1, fixed obstacle.} The obstacle is fixed at $o = [0.5, 0.0, 0.75, 0.1]$ and the target at $e = [0.399, 0.218, 0.656]$. FixedBarrierNet is compared with a fixed-obstacle classifier of identical depth. To isolate the safety mechanism, GECO weight scheduling is disabled for both, with learning rate 0.03 and $\lambda_{\mathrm{prior}} = 0.01$.
    \item \textbf{S2, multiple positions.} MultiPosBarrierNet must handle four planar obstacle positions: $(0.4, -0.2)$, $(0.5, 0.0)$, $(0.6, 0.2)$, and $(0.5, 0.3)$.
    \item \textbf{S3, general planning.} BarrierNet is evaluated on 500 randomized scenarios that match the conditions of the baseline experiment.
\end{itemize}

\subsection{Ablations and Metrics}

For every CBF variant, $\alpha$ is swept from 0.00 to 1.00 in increments of 0.05 (21 values). For the classifier, a $5 \times 7$ grid over $\lambda_{\mathrm{prior}}$ and $\lambda_{\mathrm{collision}}$ (35 configurations) ensures that the CBF is compared with the best classifier configuration. Each configuration is evaluated on 500 scenarios. We report the \emph{success rate} (target reached within 10\,mm with no collision), the \emph{goal rate} (target reached, ignoring collisions), and the \emph{collision-free (CF) rate} (no collision, ignoring convergence). Separating goal and CF rates distinguishes failures caused by over-conservatism from failures caused by insufficient safety enforcement. For the CBF we also report the \emph{intervention rate}, the fraction of optimization steps in which the corrector is applied, and the mean planning time per query. Hyperparameters are listed in Appendix~\ref{app:implementation}.

\section{Results}\label{sec:results}

\subsection{Baseline Reproduction}

Evaluated on 1{,}000 random scenarios, the reproduced baseline achieves a 70.5\% success rate (Table~\ref{tab:baseline}). A further 9.3\% of scenarios are collision-free but fail to converge, and 8.7\% reach the goal but clip an obstacle. The bottleneck is gradient quality near obstacle boundaries: the classifier's gradient is imperceptible far from an obstacle and dominant upon entry, which traps the optimizer in local minima.

\begin{table}[t]
    \centering
    \caption{Reproduced LSPP baseline (1{,}000 scenarios)}
    \label{tab:baseline}
    \begin{tabular}{cccc}
        \toprule
        Success (\%) & Goal (\%) & CF (\%) & Time (ms) \\
        \midrule
        70.5 & 79.2 & 79.8 & 186.3 \\
        \bottomrule
    \end{tabular}
\end{table}

\begin{table*}[t]
    \centering
    \caption{Decay-rate ablation for the fixed-obstacle (S1) and multi-position (S2) barriers. Best success rates in bold.}
    \label{tab:alpha_ablation}
    \begin{tabular}{c ccccc ccc}
        \toprule
        & \multicolumn{5}{c}{FixedBarrierNet $B(z)$, S1} & \multicolumn{3}{c}{MultiPosBarrierNet $B(z, x, y)$, S2} \\
        \cmidrule(lr){2-6} \cmidrule(lr){7-9}
        $\alpha$ & Success (\%) & Goal (\%) & CF (\%) & Interv. (\%) & Time (ms) & Success (\%) & Goal (\%) & CF (\%) \\
        \midrule
        0.00 & 46.8 & 56.4 & 88.2 & 90.8 & 180.8 & 36.6 & 45.2 & 90.8 \\
        0.05 & 74.4 & 86.2 & 87.0 & 67.0 & 117.4 & 79.4 & 88.6 & 90.2 \\
        0.10 & 74.4 & 88.2 & 85.0 & 52.3 & 99.6 & \textbf{83.8} & 93.6 & 89.8 \\
        0.15 & 75.0 & 90.0 & 83.2 & 43.5 & 95.6 & 83.0 & 94.8 & 87.6 \\
        0.20 & \textbf{75.6} & 91.4 & 82.0 & 37.9 & 94.8 & 81.6 & 95.0 & 85.2 \\
        0.25 & 74.0 & 92.2 & 79.2 & 34.1 & 92.3 & 79.6 & 95.8 & 82.8 \\
        0.30 & 71.0 & 92.2 & 75.8 & 31.7 & 86.8 & 79.6 & 96.0 & 82.4 \\
        1.00 & 58.0 & 99.0 & 58.4 & 19.7 & 117.0 & 56.6 & 98.4 & 57.4 \\
        \bottomrule
    \end{tabular}
\end{table*}

\subsection{Fixed Obstacle: CBF versus Classifier}

Table~\ref{tab:alpha_ablation} shows a non-monotonic success curve. At $\alpha = 0$ the corrector intervenes in 90.8\% of steps and nearly freezes the optimizer: the goal rate collapses to 56.4\% and success to 46.8\%. As $\alpha$ grows, the goal rate rises steadily to 99.0\% at $\alpha = 1$, while the CF rate falls as aggressive steps break the first-order approximation. The best trade-off occurs at $\alpha = 0.20$, with 75.6\% success (91.4\% goal, 82.0\% CF). Here the corrector is active in 37.9\% of steps, often enough to enforce safety yet sparsely enough for the goal-directed optimizer to make progress.

Table~\ref{tab:cbf_vs_classifier} compares the best CBF settings with the best classifier settings. Across all 35 grid configurations, the classifier peaks at 63.6\% ($\lambda_c = 5.0$, $\lambda_{\mathrm{prior}} = 0.01$), so the CBF improves success by 12.0 percentage points under identical base conditions (Q1). The classifier reaches its best result only with heavy weighting that suppresses goal seeking (79.4\% goal rate), whereas the predictor--corrector structure of the CBF avoids this gradient conflict.

\begin{table}[t]
    \centering
    \caption{Best CBF versus best classifier configurations (S1)}
    \label{tab:cbf_vs_classifier}
    \footnotesize
    \setlength{\tabcolsep}{4pt}
    \begin{tabular}{lcccc}
        \toprule
        Method & Success (\%) & Goal (\%) & CF (\%) & Time (ms) \\
        \midrule
        CBF, $\alpha = 0.20$ & \textbf{75.6} & 91.4 & 82.0 & 94.8 \\
        CBF, $\alpha = 0.15$ & 75.0 & 90.0 & 83.2 & 95.6 \\
        CBF, $\alpha = 0.10$ & 74.4 & 88.2 & 85.0 & 99.6 \\
        \midrule
        Classifier, $\lambda_c = 5.0$ & 63.6 & 79.4 & 68.2 & 92.9 \\
        Classifier, $\lambda_c = 10.0$ & 63.2 & 78.2 & 70.0 & 110.1 \\
        Classifier, $\lambda_c = 0.5$ & 58.8 & 91.0 & 60.4 & 101.7 \\
        \bottomrule
    \end{tabular}
\end{table}

\subsection{Multi-Position Generalization}

Conditioning the barrier on the obstacle position yields the best result overall: 83.8\% success at $\alpha = 0.10$ (93.6\% goal, 89.8\% CF), exceeding the fixed-obstacle barrier (Table~\ref{tab:alpha_ablation}). We attribute this to an implicit regularization effect: interpolating across obstacle positions forces the network to learn a smoother barrier surface with a richer gradient field than fitting a single configuration does. Even modest spatial conditioning can therefore improve the learned barrier.

\subsection{General Random-Obstacle Planning}

The fully general BarrierNet $B(z, o)$ degrades substantially, reaching at best 33.0\% success at $\alpha = 0.05$ (Table~\ref{tab:general}). We identify two causes (Q3):

\textit{Dimensionality.} The 11-D input must represent a complex boundary over random radii and heights, and 15{,}000 transition scenarios are insufficient for this generalization. Similar challenges arise in offline safe RL~\cite{zheng2024fisor} and conservative value estimation~\cite{kumar2020CQL, kostrikov2022offline}.

\textit{Covariate shift.} During planning, the average pre-correction barrier value is $-0.2786$. Because the transitions were generated by an obstacle-blind planner, the barrier routinely encounters deeply unsafe states ($B < 0$) at deployment. There, the target $(1 - \alpha \Delta t) B(z_k)$ is itself negative and the constraint no longer demands a meaningful correction: a CBF maintains forward invariance of the safe set but is not designed to recover from deep violations. This mirrors covariate shift in imitation learning~\cite{ross2011reduction, pomerleau1991efficient} and extrapolation error in offline RL~\cite{kumar2019stabilizing, levine2020OffRL}.

\begin{table}[t]
    \centering
    \caption{Decay-rate ablation for the general barrier $B(z, o)$ (S3)}
    \label{tab:general}
    \begin{tabular}{cccc}
        \toprule
        $\alpha$ & Success (\%) & Goal (\%) & CF (\%) \\
        \midrule
        0.00 & 19.2 & 40.4 & 58.4 \\
        0.05 & \textbf{33.0} & 69.4 & 39.8 \\
        0.10--1.00 & $\approx$28.0--31.0 & up to 75.8 & $\approx$30.0 \\
        \bottomrule
    \end{tabular}
\end{table}

\subsection{Computational Cost}

The corrector evaluates the barrier gradient only at steps where the constraint is violated (37.9\% of steps at $\alpha = 0.20$), so the safety filter adds little overhead. In S1, the CBF planner averages 94.8\,ms per query, on par with the best classifier (92.9\,ms) and about half the 186.3\,ms of the reproduced baseline configuration.

\section{Limitations and Future Work}\label{sec:limitations}

The main limitation is the covariate shift in the general random-obstacle task. A barrier trained on uncorrected trajectories cannot reliably enforce invariance inside an interactive optimization loop that drifts into unmapped regions of the 11-D input space, echoing distribution shift in offline RL~\cite{levine2020OffRL, fujimoto2019off} and imitation learning~\cite{ho2016generative}. In addition, the first-order correction assumes local linearity, but the VAE decoder is highly non-linear, so a step that is safe in latent space can occasionally decode into an unsafe motion in configuration space. These findings suggest three directions:

\textit{Iterative data aggregation.} Generating transitions with the CBF-corrected planner itself and feeding them back into training, in the spirit of DAgger~\cite{ross2011reduction}, directly targets the covariate shift; related ideas have succeeded in imitation~\cite{pomerleau1991efficient, ho2016generative} and goal-conditioned learning~\cite{ding2020gcil}.

\textit{Latent model predictive control.} The current planner produces open-loop trajectories. A closed-loop MPC formulation~\cite{MAYNE2000789} that operates sequentially in the latent space would let the barrier react to decoding errors at execution time.

\textit{Generative safety-aware planning.} Diffusion-based planning~\cite{janner2022planning} and feasibility-guided diffusion for offline safe RL~\cite{zheng2024fisor} could combine expressive generative models with CBF constraints. Reachability-based flow policies~\cite{tayal2026safe}, epigraph-guided flow matching for trading off safety and performance~\cite{tayal2026epiflow}, and safety filters learned from offline data~\cite{tayal2025vocbf} are particularly relevant for integrating safety-aware generative planning with latent-space methods, while robust imitation through stochastic encoding~\cite{tayal2025rise} could mitigate the distribution shift identified here. HJ reachability~\cite{bansal2017hamilton, Fisac2019HJSafety} and conformal-prediction-based neural CBFs~\cite{tayal2025cp} could further strengthen the formal safety guarantees.

\section{Conclusion}\label{sec:conclusion}

We introduced a neural control barrier function that operates directly in the latent space of a generative manipulation planner. By decoupling obstacle avoidance from goal seeking and enforcing the barrier through a closed-form, first-order correction, the method removes the cliff-edge gradients of classifier-based safety. It achieves 75.6\% and 83.8\% success on fixed-obstacle and multi-position tasks, outperforming the best-tuned classifier by 12.0 percentage points at comparable planning cost, and the decay rate $\alpha$ provides a single, interpretable knob for the safety--performance trade-off, with the best range at $[0.10, 0.20]$. The general random-obstacle setting shows that barriers trained on obstacle-blind transitions cannot recover from the unsafe states visited at deployment. Together, these results establish the feasibility of CBF-based safety constraints in learned latent spaces and lay the groundwork for closed-loop, model-predictive formulations of latent-space manipulation planning.

\appendices

\section{Implementation Details}\label{app:implementation}

All models are implemented in PyTorch and trained on a single NVIDIA GPU; hyperparameters are listed in Table~\ref{tab:hyperparameters}. The barrier networks share the classifier's architecture (four hidden layers of 2048 ELU units) but output an unbounded scalar, and their input dimension is 7, 9, or 11 depending on the variant. CBF planning uses a lower learning rate than the baseline (0.03 versus 0.15) because the correction relies on a first-order approximation that larger steps would violate, and a lower prior weight (0.01) because the correction itself limits the out-of-distribution drift that the prior term was designed to prevent. The baseline planning weights are the initial values of the GECO schedule and were varied in the classifier grid search.

\begin{table}[t]
    \centering
    \caption{Hyperparameters}
    \label{tab:hyperparameters}
    \footnotesize
    \begin{tabular}{ll}
        \toprule
        Parameter & Value \\
        \midrule
        \multicolumn{2}{l}{\textit{VAE training}} \\
        Learning rate / batch / epochs & $10^{-4}$ / 4096 / 16{,}000 \\
        GECO $g_{\mathrm{goal}}$, $g_{\mathrm{lr}}$, $g_{\alpha}$ & 0.0008, 0.005, 0.95 \\
        \midrule
        \multicolumn{2}{l}{\textit{Baseline AM planning}} \\
        Learning rate / max steps & 0.15 / 300 \\
        Initial $\lambda_{\mathrm{prior}}$ / $\lambda_{\mathrm{collision}}$ & 0.7 / 0.5 \\
        Classifier temperature $T$ & 3.0 \\
        GECO $\alpha_{\mathrm{GECO}}$ & 0.008 \\
        GECO $\tau_{\mathrm{prior}}$ / $\tau_{\mathrm{obs}}$ & 6.0 / 2.0 \\
        GECO moving averages (prior / obs.) & 0.8 / 0.8 \\
        \midrule
        \multicolumn{2}{l}{\textit{CBF training}} \\
        Learning rate & $10^{-4}$, cosine to $10^{-6}$ \\
        Batch size (label + transition) & 4096 + 4096 \\
        Margin $\gamma$ & 1.0 \\
        Loss weights $(\lambda_s, \lambda_u, \lambda_d)$ & (1.0, 1.0, 1.0) \\
        \midrule
        \multicolumn{2}{l}{\textit{CBF planning}} \\
        Learning rate / max steps & 0.03 / 300 \\
        Prior weight $\lambda_{\mathrm{prior}}$ & 0.01 \\
        Decay rate $\alpha$ / step $\Delta t$ & 0.20 / 1.0 \\
        \midrule
        Success threshold (all planners) & 10\,mm \\
        \bottomrule
    \end{tabular}
\end{table}

\bibliographystyle{IEEEtran}
\bibliography{references}

@article{hung2022reaching,
  title={Reaching Through Latent Space: From Joint Statistics to Path Planning in Manipulation},
  author={Hung, C.-M. and Ebert, F. and Mukadam, M. and Bagnell, J. A. and Levine, S.},
  journal={IEEE Robotics and Automation Letters},
  volume={7},
  number={2},
  pages={5334--5341},
  year={2022},
  publisher={IEEE}
}

@inproceedings{ichter2019robot,
  title={Robot motion planning in learned latent spaces},
  author={Ichter, B. and Pavone, M.},
  booktitle={IEEE International Conference on Robotics and Automation (ICRA)},
  pages={2440--2446},
  year={2019},
  organization={IEEE}
}

@inproceedings{robey2020learning,
  title={Learning Control Barrier Functions from Expert Demonstrations},
  author={Robey, A. and Hu, H. and Lindemann, L. and Zhang, H. and Dimarogonas, D. V. and Tu, S. and Matni, N.},
  booktitle={Proceedings of the IEEE Conference on Decision and Control (CDC)},
  pages={3710--3717},
  year={2020},
  organization={IEEE}
}

@article{ames2017control,
  title={Control barrier function based quadratic programs for safety critical systems},
  author={Ames, A. D. and Xu, X. and Grizzle, J. W. and Tabuada, P.},
  journal={IEEE Transactions on Automatic Control},
  volume={62},
  number={8},
  pages={3861--3876},
  year={2017},
  publisher={IEEE}
}

@article{srinivasan2020synthesis,
  title={Synthesis of control {Lyapunov-Barrier} functions with application to safety-critical control},
  author={Srinivasan, M. and Elbareshef, A. and Coogan, S.},
  journal={IEEE Control Systems Letters},
  volume={4},
  number={3},
  pages={572--577},
  year={2020},
  publisher={IEEE}
}

@inproceedings{kingma2014auto,
  title={Auto-encoding variational bayes},
  author={Kingma, D. P. and Welling, M.},
  booktitle={International Conference on Learning Representations (ICLR)},
  year={2014}
}

@article{rezende2018taming,
  title={Taming {VAE}s},
  author={Rezende, D. J. and Viola, F.},
  journal={arXiv preprint arXiv:1810.00597},
  year={2018}
}

@article{karaman2011sampling,
  title={Sampling-based algorithms for optimal motion planning},
  author={Karaman, S. and Frazzoli, E.},
  journal={The International Journal of Robotics Research},
  volume={30},
  number={7},
  pages={846--894},
  year={2011},
  publisher={SAGE Publications}
}

@inproceedings{ratliff2009chomp,
  title={{CHOMP}: Gradient optimization techniques for efficient motion planning},
  author={Ratliff, N. and Zucker, M. and Bagnell, J. A. and Srinivasa, S.},
  booktitle={IEEE International Conference on Robotics and Automation (ICRA)},
  pages={489--494},
  year={2009},
  organization={IEEE}
}

@inproceedings{schulman2013finding,
  title={Finding locally optimal paths with sequential convex optimization},
  author={Schulman, J. and Ho, J. and Lee, A. X. and Awwal, I. and Bradlow, H. and Abbeel, P.},
  booktitle={Robotics: Science and Systems (RSS)},
  year={2013}
}

@inproceedings{ross2011reduction,
  title={A reduction of imitation learning and structured prediction to no-regret online learning},
  author={Ross, S. and Gordon, G. and Bagnell, J. A.},
  booktitle={International Conference on Artificial Intelligence and Statistics (AISTATS)},
  year={2011}
}

@article{tayal2025rise,
  title={{RISE}: Robust Imitation through Stochastic Encoding},
  author={Tayal, Mumuksh and Tayal, Manan and Prakash, Ravi},
  journal={arXiv preprint arXiv:2503.12243},
  year={2025}
}

@article{tayal2025vocbf,
  title={{V-OCBF}: Learning Safety Filters from Offline Data via Value-Guided Offline Control Barrier Functions},
  author={Tayal, Mumuksh and Tayal, Manan and Singh, Aditya and Kolathaya, Shishir and Prakash, Ravi},
  journal={arXiv preprint arXiv:2512.10822},
  year={2025}
}

@article{tayal2026safe,
  title={Safe Flow {Q}-Learning: Offline Safe Reinforcement Learning with Reachability-Based Flow Policies},
  author={Tayal, Mumuksh and Tayal, Manan and Prakash, Ravi},
  journal={arXiv preprint arXiv:2603.15136},
  year={2026}
}

@article{tayal2026epiflow,
  title={{EpiFlow}: Bridging Safety and Performance in Offline Reinforcement Learning via Epigraph-Guided Flow Matching},
  author={Tayal, Manan and Tayal, Mumuksh and Prakash, Ravi},
  journal={arXiv preprint arXiv:2602.08054},
  year={2026}
}

@inproceedings{tayal2024control,
  title={Control barrier functions in dynamic {UAV}s for kinematic obstacle avoidance: a collision cone approach},
  author={Tayal, Manan and Singh, Rajpal and Keshavan, Jishnu and Kolathaya, Shishir},
  booktitle={2024 American Control Conference (ACC)},
  pages={3722--3727},
  year={2024},
  organization={IEEE}
}

@article{tayal2025cp,
  title={{Cp-NCBF}: A conformal prediction-based approach to synthesize verified neural control barrier functions},
  author={Tayal, Manan and Singh, Aditya and Jagtap, Pushpak and Kolathaya, Shishir},
  journal={arXiv preprint arXiv:2503.17395},
  year={2025}
}

@inproceedings{tayal2025physics,
  title={A Physics-Informed Machine Learning Framework for Safe and Optimal Control of Autonomous Systems},
  author={Tayal, Manan and Singh, Aditya and Kolathaya, Shishir and Bansal, Somil},
  booktitle={Forty-second International Conference on Machine Learning},
  year={2025}
}

@article{tayal2024semi,
  title={Semi-Supervised Safe Visuomotor Policy Synthesis using Barrier Certificates},
  author={Tayal, Manan and Singh, Aditya and Jagtap, Pushpak and Kolathaya, Shishir},
  journal={arXiv preprint arXiv:2409.12616},
  year={2024}
}

@book{sutton1998introduction,
  title={Reinforcement Learning: {A}n Introduction},
  author={Sutton, Richard S. and Barto, Andrew G.},
  publisher={The MIT Press},
  year={1998},
  address={Cambridge, MA}
}

@book{altman2021constrained,
  title={Constrained {M}arkov Decision Processes},
  author={Altman, Eitan},
  year={2021},
  publisher={Routledge}
}

@article{levine2020OffRL,
  title={Offline reinforcement learning: Tutorial, review, and perspectives on open problems},
  author={Levine, Sergey and Kumar, Aviral and Tucker, George and Fu, Justin},
  journal={arXiv preprint arXiv:2005.01643},
  year={2020}
}

@article{garcia2015comprehensive,
  title={A comprehensive survey on safe reinforcement learning},
  author={Garc{\'\i}a, Javier and Fern{\'a}ndez, Fernando},
  journal={Journal of Machine Learning Research},
  volume={16},
  number={1},
  pages={1437--1480},
  year={2015}
}

@article{Zhao2023SafeRL,
  title={Safe reinforcement learning: A survey},
  author={Zhao, Weiye and He, Tairan and Chen, Rui and Wei, Tianhao and Liu, Changliu},
  journal={arXiv preprint arXiv:2302.03122},
  year={2023}
}

@article{alshiekh2018safe,
  title={Safe Reinforcement Learning via Shielding},
  volume={32},
  journal={Proceedings of the AAAI Conference on Artificial Intelligence},
  author={Alshiekh, Mohammed and Bloem, Roderick and Ehlers, R{\"u}diger and K{\"o}nighofer, Bettina and Niekum, Scott and Topcu, Ufuk},
  year={2018}
}

@inproceedings{achiam2017constrained,
  author={Achiam, Joshua and Held, David and Tamar, Aviv and Abbeel, Pieter},
  title={Constrained policy optimization},
  year={2017},
  publisher={JMLR.org},
  booktitle={Proceedings of the 34th International Conference on Machine Learning},
  pages={22--31}
}

@inproceedings{ames2014control,
  title={Control barrier function based quadratic programs with application to adaptive cruise control},
  author={Ames, Aaron D and Grizzle, Jessy W and Tabuada, Paulo},
  booktitle={53rd IEEE Conference on Decision and Control},
  pages={6271--6278},
  year={2014},
  organization={IEEE}
}

@article{Ames2019CBF,
  title={Control barrier functions: Theory and applications},
  author={Ames, Aaron D and Coogan, Samuel and Egerstedt, Magnus and Notomista, Gennaro and Sreenath, Koushil and Tabuada, Paulo},
  journal={2019 18th European Control Conference (ECC)},
  pages={3420--3431},
  year={2019},
  organization={IEEE}
}

@inproceedings{barry2012safety,
  title={Safety verification of reactive controllers for {UAV} flight in cluttered environments using barrier certificates},
  author={Barry, Andrew J and Majumdar, Anirudha and Tedrake, Russ},
  booktitle={2012 IEEE International Conference on Robotics and Automation},
  pages={484--490},
  year={2012},
  organization={IEEE}
}

@article{nguyen2015safety,
  title={Safety-critical control for dynamical bipedal walking with precise footstep placement},
  author={Nguyen, Quan and Sreenath, Koushil},
  journal={IFAC-PapersOnLine},
  volume={48},
  number={27},
  pages={147--154},
  year={2015},
  publisher={Elsevier}
}

@inproceedings{dawson2022safe,
  title={Safe nonlinear control using robust neural {L}yapunov-barrier functions},
  author={Dawson, Charles and Qin, Zengyi and Gao, Sicun and Fan, Chuchu},
  booktitle={Conference on Robot Learning},
  pages={1724--1735},
  year={2022},
  organization={PMLR}
}

@article{dawson2023safe,
  title={Safe Control With Learned Certificates: A Survey of Neural {L}yapunov, Barrier, and Contraction Methods for Robotics and Control},
  author={Dawson, Charles and Gao, Sicun and Fan, Chuchu},
  journal={IEEE Transactions on Robotics},
  year={2023},
  publisher={IEEE}
}

@inproceedings{cosner2022end,
  title={End-to-end imitation learning with safety guarantees using control barrier functions},
  author={Cosner, Ryan K and Yue, Yisong and Ames, Aaron D},
  booktitle={2022 IEEE 61st Conference on Decision and Control (CDC)},
  pages={5316--5322},
  year={2022},
  organization={IEEE}
}

@inproceedings{NEURIPS2023_120ed726,
  author={Zhang, Hongchao and Wu, Junlin and Vorobeychik, Yevgeniy and Clark, Andrew},
  booktitle={Advances in Neural Information Processing Systems},
  title={Exact Verification of {ReLU} Neural Control Barrier Functions},
  volume={36},
  pages={5685--5705},
  year={2023}
}

@inproceedings{abate2021fossil,
  title={{FOSSIL}: A software tool for the formal synthesis of {L}yapunov functions and barrier certificates using neural networks},
  author={Abate, Alessandro and Ahmed, Daniele and Edwards, Alec and Giacobbe, Mirco and Peruffo, Andrea},
  booktitle={Proceedings of the 24th International Conference on Hybrid Systems: Computation and Control},
  pages={1--11},
  year={2021}
}

@inproceedings{so2024train,
  title={How to train your neural control barrier function: Learning safety filters for complex input-constrained systems},
  author={So, Oswin and Serlin, Zachary and Mann, Makai and Gonzales, Jake and Rutledge, Kwesi and Roy, Nicholas and Fan, Chuchu},
  booktitle={2024 IEEE International Conference on Robotics and Automation (ICRA)},
  pages={11532--11539},
  year={2024},
  organization={IEEE}
}

@InProceedings{Fernando2023iDBF,
  title={In-Distribution Barrier Functions: Self-Supervised Policy Filters that Avoid Out-of-Distribution States},
  author={Casta{\~n}eda, Fernando and Nishimura, Haruki and McAllister, Rowan Thomas and Sreenath, Koushil and Gaidon, Adrien},
  booktitle={Proceedings of The 5th Annual Learning for Dynamics and Control Conference},
  pages={286--299},
  year={2023},
  publisher={PMLR}
}

@article{ciftci2024safe,
  title={{SAFE-GIL}: {SAFEty} Guided Imitation Learning for Robotic Systems},
  author={Ciftci, Yusuf Umut and Chiu, Darren and Feng, Zeyuan and Sukhatme, Gaurav S and Bansal, Somil},
  journal={arXiv preprint arXiv:2404.05249},
  year={2024}
}

@inproceedings{bansal2017hamilton,
  title={Hamilton-{J}acobi reachability: A brief overview and recent advances},
  author={Bansal, Somil and Chen, Mo and Herbert, Sylvia and Tomlin, Claire J},
  booktitle={2017 IEEE 56th Annual Conference on Decision and Control (CDC)},
  pages={2242--2253},
  year={2017},
  organization={IEEE}
}

@inproceedings{Fisac2019HJSafety,
  author={Fisac, Jaime F. and Lugovoy, Neil F. and Rubies-Royo, Vicen{\c{c}} and Ghosh, Shromona and Tomlin, Claire J.},
  booktitle={2019 International Conference on Robotics and Automation (ICRA)},
  title={Bridging {H}amilton-{J}acobi Safety Analysis and Reinforcement Learning},
  year={2019},
  pages={8550--8556}
}

@inproceedings{kumar2020CQL,
  author={Kumar, Aviral and Zhou, Aurick and Tucker, George and Levine, Sergey},
  booktitle={Advances in Neural Information Processing Systems},
  title={Conservative {Q}-Learning for Offline Reinforcement Learning},
  volume={33},
  pages={1179--1191},
  year={2020}
}

@inproceedings{kostrikov2022offline,
  title={Offline Reinforcement Learning with Implicit {Q}-Learning},
  author={Kostrikov, Ilya and Nair, Ashvin and Levine, Sergey},
  year={2022},
  booktitle={International Conference on Learning Representations}
}

@inproceedings{fujimoto2019off,
  title={Off-policy deep reinforcement learning without exploration},
  author={Fujimoto, Scott and Meger, David and Precup, Doina},
  booktitle={International Conference on Machine Learning},
  pages={2052--2062},
  year={2019},
  organization={PMLR}
}

@article{kumar2019stabilizing,
  title={Stabilizing off-policy {Q}-learning via bootstrapping error reduction},
  author={Kumar, Aviral and Fu, Justin and Soh, Matthew and Tucker, George and Levine, Sergey},
  journal={Advances in Neural Information Processing Systems},
  volume={32},
  year={2019}
}

@inproceedings{zheng2024fisor,
  title={Safe offline reinforcement learning with feasibility-guided diffusion model},
  author={Zheng, Yinan and Li, Jianxiong and Yu, Dongjie and Yang, Yujie and Li, Shengbo Eben and Zhan, Xianyuan and Liu, Jingjing},
  booktitle={International Conference on Learning Representations},
  year={2024}
}

@article{pomerleau1991efficient,
  title={Efficient training of artificial neural networks for autonomous navigation},
  author={Pomerleau, Dean A},
  journal={Neural Computation},
  volume={3},
  number={1},
  pages={88--97},
  year={1991},
  publisher={MIT Press}
}

@article{ho2016generative,
  title={Generative adversarial imitation learning},
  author={Ho, Jonathan and Ermon, Stefano},
  journal={Advances in Neural Information Processing Systems},
  volume={29},
  year={2016}
}

@inproceedings{ng2000algorithms,
  author={Ng, Andrew Y. and Russell, Stuart J.},
  title={Algorithms for Inverse Reinforcement Learning},
  year={2000},
  booktitle={Proceedings of the Seventeenth International Conference on Machine Learning},
  pages={663--670}
}

@inproceedings{ding2020gcil,
  author={Ding, Yiming and Florensa, Carlos and Abbeel, Pieter and Phielipp, Mariano},
  booktitle={Advances in Neural Information Processing Systems},
  title={Goal-conditioned Imitation Learning},
  volume={32},
  year={2019}
}

@inproceedings{higgins2017beta,
  title={beta-{VAE}: Learning basic visual concepts with a constrained variational framework},
  author={Higgins, Irina and Matthey, Loic and Pal, Arka and Burgess, Christopher and Glorot, Xavier and Botvinick, Matthew and Mohamed, Shakir and Lerchner, Alexander},
  booktitle={International Conference on Learning Representations},
  year={2017}
}

@inproceedings{janner2022planning,
  title={Planning with Diffusion for Flexible Behavior Synthesis},
  author={Janner, Michael and Du, Yilun and Tenenbaum, Joshua and Levine, Sergey},
  booktitle={International Conference on Machine Learning},
  pages={9902--9915},
  year={2022},
  organization={PMLR}
}

@article{gallouedec2021pandagym,
  title={{panda-gym: Open-Source Goal-Conditioned Environments for Robotic Learning}},
  author={Gallou{\'e}dec, Quentin and Cazin, Nicolas and Dellandr{\'e}a, Emmanuel and Chen, Liming},
  year={2021},
  journal={4th Robot Learning Workshop: Self-Supervised and Lifelong Learning at NeurIPS}
}

@article{MAYNE2000789,
  title={Constrained model predictive control: Stability and optimality},
  journal={Automatica},
  volume={36},
  number={6},
  pages={789--814},
  year={2000},
  author={D.Q. Mayne and J.B. Rawlings and C.V. Rao and P.O.M. Scokaert}
}

@article{gottesman2019guidelines,
  title={Guidelines for reinforcement learning in healthcare},
  author={Gottesman, Omer and Johansson, Fredrik and Komorowski, Matthieu and Faisal, Aldo and Sontag, David and Doshi-Velez, Finale and Celi, Leo Anthony},
  journal={Nature Medicine},
  volume={25},
  number={1},
  pages={16--18},
  year={2019},
  publisher={Nature Publishing Group}
}

@inproceedings{wang2023enforcing,
  title={Enforcing hard constraints with soft barriers: Safe reinforcement learning in unknown stochastic environments},
  author={Wang, Yixuan and Zhan, Simon Sinong and Jiao, Ruochen and Wang, Zhilu and Jin, Wanxin and Yang, Zhuoran and Wang, Zhaoran and Huang, Chao and Zhu, Qi},
  booktitle={International Conference on Machine Learning},
  pages={36593--36604},
  year={2023},
  organization={PMLR}
}

@inproceedings{choi2021robust,
  title={Robust control barrier--value functions for safety-critical control},
  author={Choi, Jason J and Lee, Donggun and Sreenath, Koushil and Tomlin, Claire J and Herbert, Sylvia L},
  booktitle={2021 60th IEEE Conference on Decision and Control (CDC)},
  pages={6814--6821},
  year={2021},
  organization={IEEE}
}

\end{document}